\documentclass[11pt,a4paper]{article}
\usepackage[slovak]{babel}
\usepackage[utf8]{inputenc}
\usepackage[hyperref]{emnlp2018}
\usepackage{times}
\usepackage{latexsym}
\usepackage{graphicx}
\usepackage{amsmath}
\usepackage[all]{nowidow}

\usepackage{url}

\aclfinalcopy 

\title{Improving Moderation of Online Discussions via Interpretable Neural Models}

\author{Andrej Švec\textsuperscript{1}, Matúš Pikuliak\textsuperscript{2}, Marián Šimko\textsuperscript{2}, Mária Bieliková\textsuperscript{2} \\
  Slovak University of Technology in Bratislava, Bratislava, Slovakia \\
  Faculty of Informatics and Information Technologies \\
  \textsuperscript{1}{\tt andy.swec@gmail.com}\\
  \textsuperscript{2}{\tt \{matus.pikuliak,marian.simko,maria.bielikova\}@stuba.sk}
}
  
\date{}

\begin{document}
\maketitle
\begin{abstract}
Growing amount of comments make online discussions difficult to moderate by human moderators only. Antisocial behavior is a common occurrence that often discourages other users from participating in discussion. We propose a neural network based method that partially automates the moderation process. It consists of two steps. First, we detect inappropriate comments for moderators to see. Second, we highlight inappropriate parts within these comments to make the moderation faster. We evaluated our method on data from a major Slovak news discussion platform.
\end{abstract}

\section{Introduction}

Keeping the discussion on a website civil is important for user satisfaction as well as for legal reasons~\cite{echr-delfi-estonia}. Manually moderating all the comments might be too time consuming. Larger news and discussion websites receive hundreds of comments per minute which might require huge moderator teams. In addition it is easy to overlook inappropriate comments due to human error. Automated solutions are being developed to reduce moderation time requirements and to mitigate the error rate.

In this work we propose a neural network based method to speed up the moderation process. First, we use trained classifier to automatically detect inappropriate comments. Second, a subset of words is selected with a method by~\cite{lei-rationalizing-neural-predictions} based on reinforcement learning. These selected words should form a rationale why a comment was classified as inappropriate by our model. Selected words are then highlighted for moderators so they can quickly focus on problematic parts of comments. We also managed to evaluate our solution on a major dataset (millions of comments) and in real world conditions at an important Slovak news discussion platform.


\section{Related work}

\paragraph{Inappropriate comments detection.} There are various approaches to detection of inappropriate comments in online discussions \cite{schmidt-hate-speech-survey}. The most common approach is to detect inappropriate texts through machine learning. Features used include bag of words~\cite{burnap-bow}, lexicons~\cite{gitari-lexicon}, linguistic, syntactic and sentiment features~\cite{nobata-linguistic-syntactic}, Latent Dirichlet Allocation features~\cite{zhong-content-lda} or comment embeddings~\cite{djuric-comment-embeddings}. Deep learning was also considered to tackle this issue~\cite{badjatiya-deep-learning-hate-speech, mehdad-characters-deep-learning}.

Apart from detecting inappropriate texts, multiple works focus on detecting users that should be banned~\cite{cheng-antisocial-behavior} by analyzing their posts and their activity in general~\cite{adler-wikipedia-vandalism-detection, ribeiro-characterizing-detecting-users}, their relationships with other users~\cite{ribeiro-hateful-users-relationships} and the reaction of other users~\cite{cheng-community-feedback} or moderators~\cite{cheng-antisocial-behavior} towards them.

\paragraph{Interpreting neural models.} Interpretability of machine learning models is common requirement when deploying the models to production. In our case moderators would like to know why was the comment marked as inappropriate.

Most of the works deal with interpretability of computer vision models~\cite{DBLP:conf/eccv/ZeilerF14}, but progress in interpretable text processing was also made.
Several works try to analyze the dynamics of what is happening inside the neural network. \citet{karpathy-visualizing-rnn, D16-1216} focus on memory cells activations. \citet{li-visualizing-neural-models} compute how much individual input units contribute to the final decision. Other techniques rely on attention mechanisms~\cite{DBLP:conf/naacl/YangYDHSH16}, contextual decomposition~\cite{murdoch-iclr}, representation erasure~\cite{DBLP:journals/corr/LiMJ16a} or relevance propagation~\cite{10.1371/journal.pone.0181142}. Our work uses the method by ~\citet{lei-rationalizing-neural-predictions}, which selects coherent subset of words responsible for neural network decision. The authors use this model to explain multi-aspect sentiment analysis over beer reviews and information retrieval over CQA system.

In the domain of detecting antisocial behavior \citet{D17-1117}~used attention mechanism to interpret existing model. Their work is the most relevant to ours, but we use other datasets as well as other, more explicit, technique for model interpretation.


\section{Interpretable Neural Moderation}

We propose a method to speed up the moderation process of online discussions. It consists of two steps:

\begin{enumerate}
    \item We detect inappropriate comments. This is a binary classification problem. Comments are sorted by model confidence and shown to the moderators. In effect after this initial filtering moderators work mostly with inappropriate comments which improves their efficiency.
    \item We highlight critical parts of inappropriate comments to convince moderators that selected comments are indeed harmful. The moderators can then focus on these highlighted parts instead of reading the whole comment.
\end{enumerate}

\subsection{Step 1: Inappropriate comments detection}

We approach inappropriate comments detection as a binary classification problem. Each comment is either appropriate or inappropriate. We use recurrent neural network that takes sequence of word embeddings as input. The final output is then used to predict the probability of comment being inappropriate. We use RCNN recurrent cells~\cite{barzilay-rcnn} instead of more commonly used LSTM cells as they proved to be faster to train with practically identical results. This part of our method is trained in supervised fashion using Adam optimization algorithm.

\subsection{Step 2: Inappropriate parts highlighting}

Our method implements \cite{lei-rationalizing-neural-predictions}. It can learn to select the words responsible for a decision of a neural network called {\it rationale} without the need for word level annotations in the data.

The model processes comment word embeddings $x$ and generates two outputs: binary flags $z$ representing selection of individual words into rationale which is marked $(z, x)$ and $y$ being probability distribution over classes appropriate / inappropriate. The model is composed of two modules: generator $gen$ and classifier $clas$ called also encoder in the original work.

\paragraph{Generator $gen$.} The role of generator is to select words that are responsible for a comment being in/appropriate. On its output layer it generates probabilities of selection for each word $p(z|x)$. A well trained model assigns high probability scores to words that should form the rationale and low scores to the rest. In the final step these probabilities are used to sample binary selections $z$. The sampling layer is called Z-layer.

Due to sampling in Z-layer $gen$ graph becomes non-differentiable. To overcome this issue the method uses reinforcement learning method called policy gradients \cite{williams-policy-gradients} to train the generator.

\paragraph{Classifier $clas$.} $clas$ is a softmax classifier that tries to determine whether a comment is inappropriate or not by processing only words from rationale $(z, x)$.

\paragraph{Joint learning.} In order to learn to highlight inappropriate words from inappropriate comments we need $gen$ and $clas$ to cooperate. $gen$ selects words and $clas$ provides feedback on the quality of selected words. The feedback is based on the assumption that the words are selected correctly if $clas$ is able to classify comment correctly based on the rationale $(z, x)$ and vice versa.

Furthermore, there are some conditions on the rationale: it must to be short and meaningful (the selected words must be near each other) what is achieved by adding regularization controlled by hyperparameters $\lambda_1$ (which forces the rationales to have fewer words) and $\lambda_2$ (which forces the selected words to be in a row). The following loss function expresses these conditions:
\begin{multline}
loss(x,z,y') = \| clas(z, x) - y' \|_2^2 \\ + \lambda_1 \| z \| + \lambda_2 \sum_{t=1}^{K-1} | z_t - z_{t+1}|
\end{multline}
where $x$ is original comment text, $z$ contains binary flags representing non/selection of each word in $x$, $(z, x)$ contains actual words selected to rationale, $y'$ is correct output and $K$ is the length of $x$ and also $z$ respectively.

From the loss function we can see that the training is based on a simple assumption that rationale is a subset of words that $clas$ classifies correctly. If it is not classified correctly then the rationale is probably incorrect. This way we can learn to generate rationales without need to have word level annotations. We would like to make the point that training to generate these rationales is not done to improve the classification performance. It uses only the exact same data the classifier from Step 1 uses. Its only effect is to generate interpretable rationales behind the decisions classifier takes.


\section{Experiments and Results}

\subsection{Dataset}

We used a proprietary dataset of more than 20 million comments from a major Slovak news discussion platform. Over the years a team of moderators was considering reported comments and removing the inappropriate ones while also selecting a reason(s) from prepared list of possible discussion code violations. In this work we consider only those that were flagged because of following reasons: insults, racism, profanity or spam. The rest of the comments are considered appropriate. We split the dataset in train, validation and test set where validation and test set both were balanced to contain 10,000 appropriate and 10,000 inappropriate comments. Rest of the dataset forms the training set. Test and validation sets were sampled from the most recent months. During the training we balance it on batch level by supersampling inappropriate comments.

\paragraph{Highlights test set.} We did not have any annotations on rationales in the dataset. We created a test set by manually selecting words that should form the rationales in randomly picked 100~comments. This way we created a test set containing 3,600 annotated words.

\paragraph{Word embeddings.} We trained our own fastText embeddings~\cite{bojanowski-vectors-subword} on our dataset. These take into account character level information and are therefore suitable for inflected languages (such as Slovak) and online discussions where lots of grammatical and typing errors occur.

\subsection{Inappropriate comments detection}

We performed a hyperparameter grid search with our method. We experimented with different recurrent cells (RCNN and LSTM), depth (2, 3), hidden size (200, 300, 500), bi-directional RNN and in the case of RCNN also with cell order (2, 4). We also trained several non-neural methods for comparison. Results from this experiment are marked in Table~\ref{tab:classification-results}. We measure accuracy as well as average precision (AP). The best results were achieved by bi-directional 2-layer RCNN with hidden size 300 and order 2. Deep neural network models outperform feature based models by almost 10\% of accuracy. RCNN achieves results similar to LSTM but with approximately 8.5~times less parameters.

The results here might be significantly affected by noisy data. During the years many inappropriate comments went unnoticed and many appropriate comments were blocked if they were in inappropriate threads. Qualitative interviews we carried out with moderators indicate that our accuracy might be a bit higher.

We observed that model was the most confident about insulting and offensive comments. Thanks to sub-word based word embeddings the model can find profanities even when some characters within are replaced with numbers (e.g. \emph{1nsult} instead of insult) or there are arbitrary characters inserted into the word (e.g. \emph{i..n..s..u..l..t} instead of \emph{insult}.

\begin{figure}[tb]
    \begin{center}
        \includegraphics[width=\linewidth]{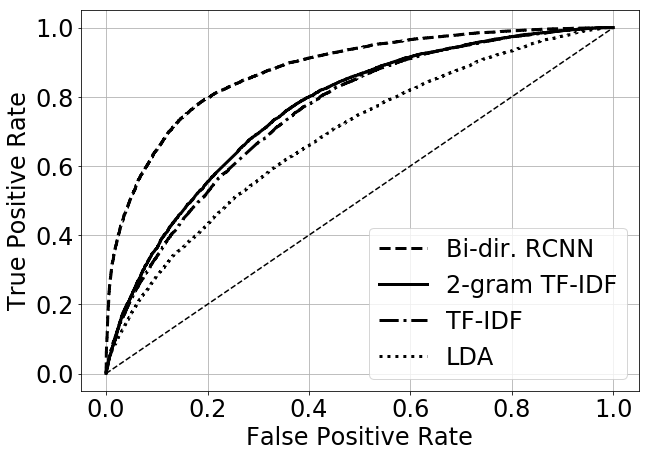}
        \caption{Receiver operating characteristic (ROC) of different models. We plot only one neural based solution as they almost completely overlap.}
        \label{fig:roc}
    \end{center}
\end{figure}

To better understand the impact of this classifier we plotted its results using ROC curve in Figure~\ref{fig:roc}. Here we can see how many comments a moderator needs to read to find certain percentile of inappropriate ones. E.g. when looking for 80\% of inappropriate comments, only 20\% of reviewed comments will be falsely flagged by the model.

\begin{table}[tb]
    \centering
    \begin{tabular}{|l|ccc|}
        \hline
        Model & \# params & \% acc & AUC \\
        \hline
        LDA (75 topics) & - & 63.2 & 0.684 \\
        LSA (300 topics) & - & 66.2 & - \\
        TF-IDF & - & 68.8 & 0.754 \\
        2-gram TF-IDF & - & 70.1 & 0.766 \\
        Unidir. RCNN & 0.6M & 79.0 & 0.870 \\
        Unidir. LSTM & 3.3M & 78.9 & 0.872 \\
        \textbf{Bi-dir. RCNN} & \textbf{1.7M} & \textbf{79.4} & \textbf{0.872} \\
        \textbf{Bi-dir. LSTM} & \textbf{14.6M} & \textbf{79.4} & \textbf{0.875} \\
        \hline
    \end{tabular}
    \caption{Comparison of test set performance of multiple classification models. Baseline models (first four) use various text representations that are classified by boosted decision trees (1,000 trees)~\cite{freund-adaboost}.}
    \label{tab:classification-results}
\end{table}

\subsection{Highlighting inappropriate parts}
\label{experiments-results:highlighting-inappropriate-parts}

$gen$ and $clas$ are implemented as recurrent neural networks with RCNN cells. $gen$ is a bi-directional 2-layer RCNN with hidden size equal 200 and order equal 2. Z-layer is realized as unidirectional RNN with hidden size equal 30. $clas$ is an unidirectional 2-layer RCNN with hidden size equal 200 and order equal 2. For regularization hyperparameters we found values $\lambda_1 \in [5\times10^{-4},3\times10^{-3}]$ and $\lambda_2 \in [2\lambda_1, 4\lambda_1]$ to perform well.

We observed a significant instability during the training caused by formulation of our loss function. The model would often converge to a state where it would pick all the words or no words at all. Especially the cases when the model started to pick all the words proved to be impossible to overcome. In such cases we restarted the training from a different seed what increased the probability of a model converging successfully by a factor of five.

We evaluated following metrics of our models:

\begin{itemize}
    \item {\it Precision} -- how many of selected words were actually part of golden inappropriate data. Correct selection of words is a prerequisite for saving moderators' time. Recall is not very important as we do not need to select all the inappropriate parts. One part is usually enough for the moderators to block a comment.
    \item {\it Rationale length} -- the proportion of words selected into rationale. It is important to measure this metric as we want our model to only pick a handful of strongly predictive words.
\end{itemize}

\begin{figure}[tb]
    \begin{center}
        \includegraphics[width=\linewidth]{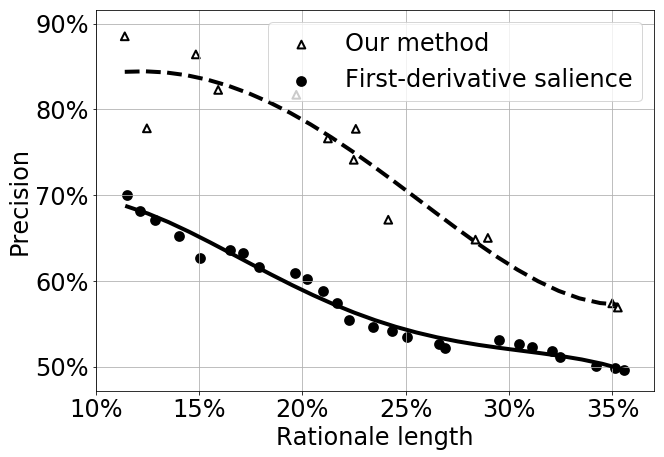}
        \caption{Test set performance of models highlighting inappropriate parts in comments.}
        \label{fig:highlight-results}
    \end{center}
\end{figure}

We compare our proposed model with the model based on first-derivative saliency~\cite{li-visualizing-neural-models}. The comparison of models is shown in Figure~\ref{fig:highlight-results}. We can see that with length reduction the precision grows as expected. Our best models achieve a precision of nearly 90\% while selecting 10--15\% of words. We consider this to be very good result. Our method outperforms the saliency based one and also produces less scattered rationales. By this we mean that the average length of a segment of subsequently selected words is 2.5 for our method, but only 1.5 for saliency-based method. Instead of picking individual words our model tries to pick longer segments.

\section{Conclusion}

Moderating online discussions is time consuming error-prone activity. Major discussion platforms have millions of users so they need huge teams of moderators. We propose a method to speed up this process and make it more reliable. The novelty of our approach is in the application of a model interpretation method in this domain.

Instead of simply marking the comment as inappropriate, our method highlights the words that made the model think so. This is significant help for moderators as they can now read only small part of comment instead of its whole text. We believe that our method can significantly speed up the moderation process and user study is underway to confirm this hypothesis.

We evaluated our model on dataset from a major Slovak news discussion platform with more than 20 million comments. Results are encouraging as we were able to obtain good results on inappropriate comments detection task. We also obtained good results (nearly 90\% precision) when highlighting inappropriate parts of these comments.

In the future we plan to improve the evaluation of highlighting. Instead of measuring the global precision, we plan to analyze how well it performs with various types of inappropriateness, such as racism, insults or spam. We are also looking into the possibility of incorporating additional data into our algorithm -- other comments from the same thread, article for which the comments are created or even user profiles. These could help us improve our results or even detect the possibility of antisocial behavior before it even happens.

\section*{Acknowledgments}
This work was partially supported by the Slovak Research and Development Agency under the contracts No. APVV-17-0267 - Automated Recognition of Antisocial Behaviour in Online Communities and No. APVV-15-0508 - Human Information Behavior in the Digital Space. The authors would like to thank for financial contribution from the Scientific Grant Agency of the Slovak Republic, grant No. VG 1/0646/15.     

\bibliography{main}

\begin{thebibliography}{28}
\expandafter\ifx\csname natexlab\endcsname\relax\def\natexlab#1{#1}\fi

\bibitem[{Adler et~al.(2011)Adler, De~Alfaro, Mola-Velasco, Rosso, and
  West}]{adler-wikipedia-vandalism-detection}
B~Thomas Adler, Luca De~Alfaro, Santiago~M Mola-Velasco, Paolo Rosso, and
  Andrew~G West. 2011.
\newblock Wikipedia vandalism detection: Combining natural language, metadata,
  and reputation features.
\newblock In \emph{International Conference on Intelligent Text Processing and
  Computational Linguistics}, pages 277--288. Springer.

\bibitem[{Arras et~al.(2017)Arras, Horn, Montavon, Müller, and
  Samek}]{10.1371/journal.pone.0181142}
Leila Arras, Franziska Horn, Grégoire Montavon, Klaus-Robert Müller, and
  Wojciech Samek. 2017.
\newblock "what is relevant in a text document?": An interpretable machine
  learning approach.
\newblock \emph{PLOS ONE}, 12(8):1--23.

\bibitem[{Aubakirova and Bansal(2016)}]{D16-1216}
Malika Aubakirova and Mohit Bansal. 2016.
\newblock Interpreting neural networks to improve politeness comprehension.
\newblock In \emph{Proceedings of the 2016 Conference on Empirical Methods in
  Natural Language Processing}, pages 2035--2041. Association for Computational
  Linguistics.

\bibitem[{Badjatiya et~al.(2017)Badjatiya, Gupta, Gupta, and
  Varma}]{badjatiya-deep-learning-hate-speech}
Pinkesh Badjatiya, Shashank Gupta, Manish Gupta, and Vasudeva Varma. 2017.
\newblock Deep learning for hate speech detection in tweets.
\newblock In \emph{Proceedings of the 26th International Conference on World
  Wide Web Companion}, pages 759--760. International World Wide Web Conferences
  Steering Committee.

\bibitem[{Barzilay et~al.(2016)Barzilay, Jaakkola, Tymoshenko, and
  Marquez}]{barzilay-rcnn}
Tao Lei Hrishikesh Joshi~Regina Barzilay, Tommi Jaakkola, Katerina Tymoshenko,
  and Alessandro Moschitti~Llu{\i}s Marquez. 2016.
\newblock Semi-supervised question retrieval with gated convolutions.
\newblock In \emph{Proceedings of NAACL-HLT}, pages 1279--1289.

\bibitem[{Bojanowski et~al.(2017)Bojanowski, Grave, Joulin, and
  Mikolov}]{bojanowski-vectors-subword}
Piotr Bojanowski, Edouard Grave, Armand Joulin, and Tomas Mikolov. 2017.
\newblock Enriching word vectors with subword information.
\newblock \emph{Transactions of the Association of Computational Linguistics},
  5(1):135--146.

\bibitem[{Burnap and Williams(2016)}]{burnap-bow}
Pete Burnap and Matthew~L Williams. 2016.
\newblock Us and them: identifying cyber hate on twitter across multiple
  protected characteristics.
\newblock \emph{EPJ Data Science}, 5(1):11.

\bibitem[{Cheng et~al.(2014)Cheng, Danescu-Niculescu-Mizil, and
  Leskovec}]{cheng-community-feedback}
Justin Cheng, Cristian Danescu-Niculescu-Mizil, and Jure Leskovec. 2014.
\newblock How community feedback shapes user behavior.
\newblock In \emph{Eighth International AAAI Conference on Weblogs and Social
  Media}.

\bibitem[{Cheng et~al.(2015)Cheng, Danescu-Niculescu-Mizil, and
  Leskovec}]{cheng-antisocial-behavior}
Justin Cheng, Cristian Danescu-Niculescu-Mizil, and Jure Leskovec. 2015.
\newblock Antisocial behavior in online discussion communities.
\newblock In \emph{Ninth International AAAI Conference on Web and Social
  Media}.

\bibitem[{Djuric et~al.(2015)Djuric, Zhou, Morris, Grbovic, Radosavljevic, and
  Bhamidipati}]{djuric-comment-embeddings}
Nemanja Djuric, Jing Zhou, Robin Morris, Mihajlo Grbovic, Vladan Radosavljevic,
  and Narayan Bhamidipati. 2015.
\newblock Hate speech detection with comment embeddings.
\newblock In \emph{Proceedings of the 24th international conference on world
  wide web}, pages 29--30. ACM.

\bibitem[{{European court of human rights}(2015)}]{echr-delfi-estonia}
{European court of human rights}. 2015.
\newblock {Case of Delfi as v. Estonia}.
\newblock \url{https://hudoc.echr.coe.int/eng#{"itemid":["001-155105"]}}.
\newblock [Online; accessed February 23th, 2018].

\bibitem[{Freund and Schapire(1995)}]{freund-adaboost}
Yoav Freund and Robert~E Schapire. 1995.
\newblock A desicion-theoretic generalization of on-line learning and an
  application to boosting.
\newblock In \emph{European conference on computational learning theory}, pages
  23--37. Springer.

\bibitem[{Gitari et~al.(2015)Gitari, Zuping, Damien, and Long}]{gitari-lexicon}
Njagi~Dennis Gitari, Zhang Zuping, Hanyurwimfura Damien, and Jun Long. 2015.
\newblock A lexicon-based approach for hate speech detection.
\newblock \emph{International Journal of Multimedia and Ubiquitous
  Engineering}, 10(4):215--230.

\bibitem[{Karpathy et~al.(2015)Karpathy, Johnson, and
  Fei-Fei}]{karpathy-visualizing-rnn}
Andrej Karpathy, Justin Johnson, and Li~Fei-Fei. 2015.
\newblock Visualizing and understanding recurrent networks.
\newblock \emph{arXiv preprint arXiv:1506.02078}.

\bibitem[{Lei et~al.(2016)Lei, Barzilay, and
  Jaakkola}]{lei-rationalizing-neural-predictions}
Tao Lei, Regina Barzilay, and Tommi Jaakkola. 2016.
\newblock Rationalizing neural predictions.
\newblock In \emph{Proceedings of the 2016 Conference on Empirical Methods in
  Natural Language Processing}, pages 107--117.

\bibitem[{Li et~al.(2016{\natexlab{a}})Li, Chen, Hovy, and
  Jurafsky}]{li-visualizing-neural-models}
Jiwei Li, Xinlei Chen, Eduard Hovy, and Dan Jurafsky. 2016{\natexlab{a}}.
\newblock Visualizing and understanding neural models in nlp.
\newblock In \emph{Proceedings of NAACL-HLT}, pages 681--691.

\bibitem[{Li et~al.(2016{\natexlab{b}})Li, Monroe, and
  Jurafsky}]{DBLP:journals/corr/LiMJ16a}
Jiwei Li, Will Monroe, and Dan Jurafsky. 2016{\natexlab{b}}.
\newblock Understanding neural networks through representation erasure.
\newblock \emph{CoRR}, abs/1612.08220.

\bibitem[{Mehdad and Tetreault(2016)}]{mehdad-characters-deep-learning}
Yashar Mehdad and Joel Tetreault. 2016.
\newblock Do characters abuse more than words?
\newblock In \emph{Proceedings of the 17th Annual Meeting of the Special
  Interest Group on Discourse and Dialogue}, pages 299--303.

\bibitem[{Murdoch et~al.(2018)Murdoch, Liu, and Yu}]{murdoch-iclr}
W.~James Murdoch, Peter~J. Liu, and Bin Yu. 2018.
\newblock Beyond word importance: Contextual decomposition to extract
  interactions from lstms.
\newblock In \emph{Proceedings of International Conference on Learning
  Representations (ICLR)}.

\bibitem[{Nobata et~al.(2016)Nobata, Tetreault, Thomas, Mehdad, and
  Chang}]{nobata-linguistic-syntactic}
Chikashi Nobata, Joel Tetreault, Achint Thomas, Yashar Mehdad, and Yi~Chang.
  2016.
\newblock Abusive language detection in online user content.
\newblock In \emph{Proceedings of the 25th international conference on world
  wide web}, pages 145--153. International World Wide Web Conferences Steering
  Committee.

\bibitem[{Pavlopoulos et~al.(2017)Pavlopoulos, Malakasiotis, and
  Androutsopoulos}]{D17-1117}
John Pavlopoulos, Prodromos Malakasiotis, and Ion Androutsopoulos. 2017.
\newblock Deeper attention to abusive user content moderation.
\newblock In \emph{Proceedings of the 2017 Conference on Empirical Methods in
  Natural Language Processing}, pages 1125--1135. Association for Computational
  Linguistics.

\bibitem[{Ribeiro et~al.(2018{\natexlab{a}})Ribeiro, Calais, Santos, Almeida,
  and Meira~Jr}]{ribeiro-characterizing-detecting-users}
Manoel~Horta Ribeiro, Pedro~H Calais, Yuri~A Santos, Virg{\'\i}lio~AF Almeida,
  and Wagner Meira~Jr. 2018{\natexlab{a}}.
\newblock Characterizing and detecting hateful users on twitter.
\newblock \emph{arXiv preprint arXiv:1803.08977}.

\bibitem[{Ribeiro et~al.(2018{\natexlab{b}})Ribeiro, Calais, Santos, Almeida,
  and Meira~Jr}]{ribeiro-hateful-users-relationships}
Manoel~Horta Ribeiro, Pedro~H Calais, Yuri~A Santos, Virg{\'\i}lio~AF Almeida,
  and Wagner Meira~Jr. 2018{\natexlab{b}}.
\newblock “like sheep among wolves”: Characterizing hateful users on
  twitter.

\bibitem[{Schmidt and Wiegand(2017)}]{schmidt-hate-speech-survey}
Anna Schmidt and Michael Wiegand. 2017.
\newblock A survey on hate speech detection using natural language processing.
\newblock In \emph{Proceedings of the Fifth International Workshop on Natural
  Language Processing for Social Media}, pages 1--10.

\bibitem[{Williams(1992)}]{williams-policy-gradients}
Ronald~J Williams. 1992.
\newblock Simple statistical gradient-following algorithms for connectionist
  reinforcement learning.
\newblock In \emph{Reinforcement Learning}, pages 5--32. Springer.

\bibitem[{Yang et~al.(2016)Yang, Yang, Dyer, He, Smola, and
  Hovy}]{DBLP:conf/naacl/YangYDHSH16}
Zichao Yang, Diyi Yang, Chris Dyer, Xiaodong He, Alexander~J. Smola, and
  Eduard~H. Hovy. 2016.
\newblock Hierarchical attention networks for document classification.
\newblock In \emph{{NAACL} {HLT} 2016, The 2016 Conference of the North
  American Chapter of the Association for Computational Linguistics: Human
  Language Technologies, San Diego California, USA, June 12-17, 2016}, pages
  1480--1489.

\bibitem[{Zeiler and Fergus(2014)}]{DBLP:conf/eccv/ZeilerF14}
Matthew~D. Zeiler and Rob Fergus. 2014.
\newblock Visualizing and understanding convolutional networks.
\newblock In \emph{Computer Vision - {ECCV} 2014 - 13th European Conference,
  Zurich, Switzerland, September 6-12, 2014, Proceedings, Part {I}}, pages
  818--833.

\bibitem[{Zhong et~al.(2016)Zhong, Li, Squicciarini, Rajtmajer, Griffin,
  Miller, and Caragea}]{zhong-content-lda}
Haoti Zhong, Hao Li, Anna Squicciarini, Sarah Rajtmajer, Christopher Griffin,
  David Miller, and Cornelia Caragea. 2016.
\newblock Content-driven detection of cyberbullying on the instagram social
  network.
\newblock In \emph{Proceedings of the Twenty-Fifth International Joint
  Conference on Artificial Intelligence}, pages 3952--3958. AAAI Press.

\end{thebibliography}
\bibliographystyle{acl_natbib_nourl}


\end{document}